\documentclass[10pt,twocolumn,letterpaper]{article}

\usepackage{cvpr}              
\usepackage{adjustbox}
\definecolor{cvprblue}{rgb}{0.21,0.49,0.74}
\usepackage[pagebackref,breaklinks,colorlinks,allcolors=cvprblue]{hyperref}

\def\paperID{43767} 
\def\confName{CVPR}
\def\confYear{2026}

\title{U-Mind: A Unified Framework for Real-Time Multimodal Interaction with Audiovisual Generation}

\author{
    Xiang Deng$^{1}$, 
    Feng Gao$^{2}$, 
    Yong Zhang$^{2}$, 
    Youxin Pang$^{1}$, 
    Xu Xiaoming$^{2}$, \\
    Zhuoliang Kang$^{2}$, 
    Xiaoming Wei$^{2}$, 
    Yebin Liu$^{1}$ \\
    $^{1}$Tsinghua University \quad $^{2}$Meituan \\
}
\begin{document}
\twocolumn[{%
\renewcommand\twocolumn[1][]{#1}%
\maketitle
\begin{center}
    \centering
    \captionsetup{type=figure}
    \includegraphics[width=0.9\textwidth]{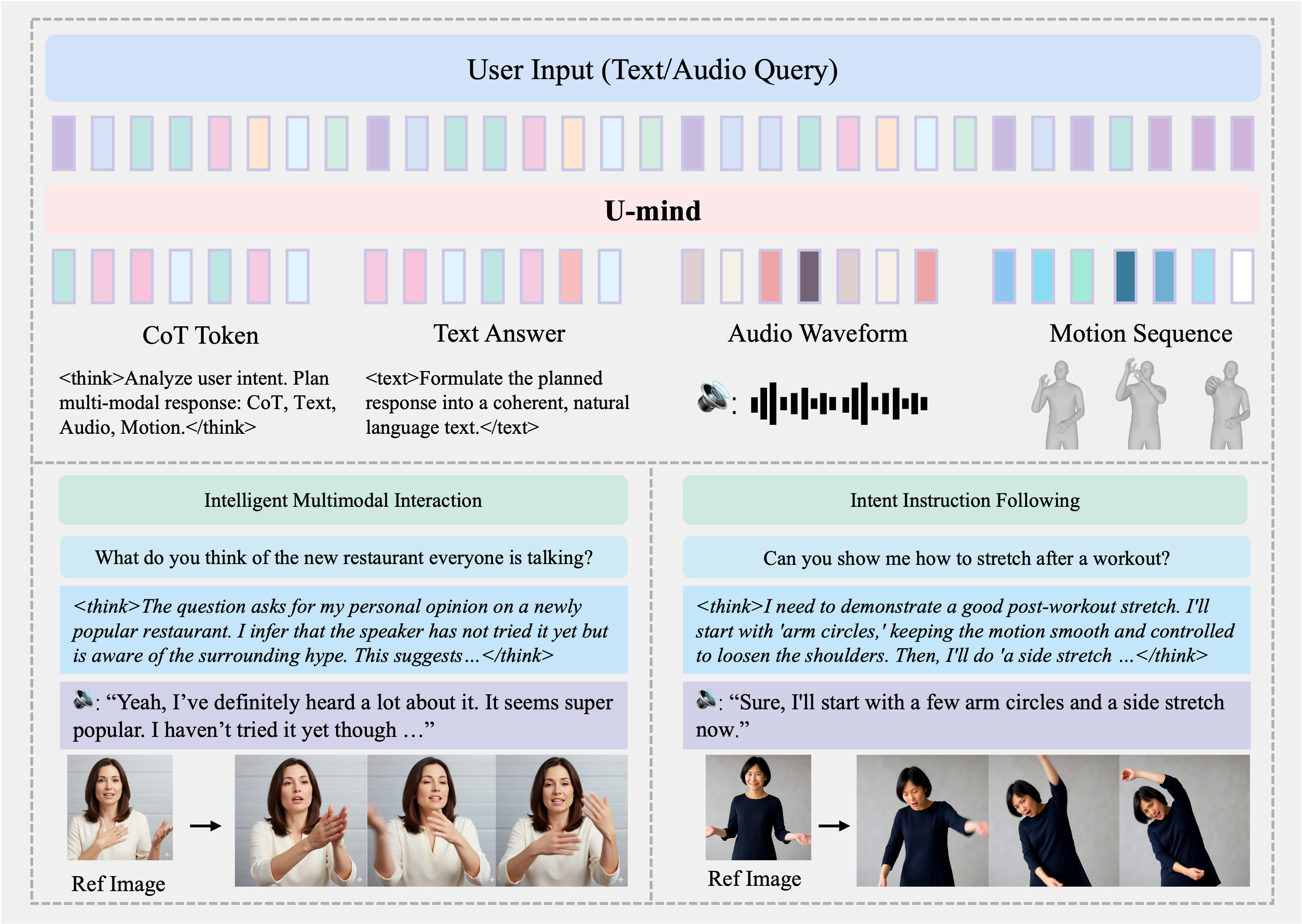}
    \captionof{figure}{Given a user query in text or speech, our system performs internal Chain-of-Thought (CoT) planning and produces synchronized responses across text, speech, and gesture. As shown, the model can handle both open-domain dialogue and various instruction-following, generating coherent language, natural prosody, and expressive body motion. The final output is rendered into photorealistic talking videos, showcasing our framework’s capability for high-level multimodal understanding and generation.}
    \label{fig:teaser}
\end{center}%
}]
\begin{abstract}

Full-stack multimodal interaction in real-time is a central goal in building intelligent embodied agents capable of natural, dynamic communication. However, existing systems are either limited to unimodal generation or suffer from degraded reasoning and poor cross-modal alignment, preventing coherent and perceptually grounded interactions. 
In this work, we introduce \textbf{U-Mind}, the first unified system for high-intelligence multimodal dialogue that supports real-time generation and jointly models language, speech, motion, and video synthesis within a single interactive loop.
At its core, U-Mind implements a \textit{Unified Alignment and Reasoning Framework} that addresses two key challenges: enhancing cross-modal synchronization via a \textit{segment-wise alignment strategy}, and preserving reasoning abilities through \textit{Rehearsal-Driven Learning}. 
During inference, U-Mind adopts a \textit{text-first decoding pipeline} that performs internal chain-of-thought planning followed by temporally synchronized generation across modalities. To close the loop, we implement a real-time video rendering framework conditioned on pose and speech, enabling expressive and synchronized visual feedback.
Extensive experiments demonstrate that U-Mind achieves state-of-the-art performance on a range of multimodal interaction tasks, including question answering, instruction following, and motion generation, paving the way toward intelligent, immersive conversational agents.

\end{abstract}    
\section{Introduction}
\label{sec:intro}
The creation of interactive digital humans capable of real-time, multimodal, and closed-loop interaction represents a pivotal milestone in embodied artificial intelligence. While prior systems have achieved fluent dialogue through text or speech, they lack visual interaction capabilities crucial for grounded, embodied communication \cite{shao2023character,zhang2023speechgpt,xu2025qwen3,fang2025llama}. Other efforts have explored gesture or motion generation, but often treat it as a one-way mapping from text or audio input, without high-level reasoning ~\cite{liu2022beat,yi2023generating,kucherenko2020gesticulator,yoon2019robots,ahuja2019language2pose,danvevcek2023emotional,liu2022learning, pang2023bodyformer,qi2023weakly,zhu2023taming,karunratanakul2023guided,tevethuman}. 
Consequently, these models struggle to adapt to dynamic contexts or perform complex, goal-directed interactions.

The core challenge in building an intelligent, fully interactive agent lies in generating coherent multimodal outputs that combine high-level reasoning with expressive body motion generation. To address this, recent works have introduced large language models (LLMs) as external planners to enhance semantic control in motion generation~\cite{wu2025motion,han2025atom,zhang2024semantic,cong2025semgeomo,tan2025think}.
Other approaches integrate motion generation into large language models by discretizing body movements into motion tokens, enabling autoregressive, end-to-end generation of language and motion within a shared token space~\cite{jiang2023motiongpt,wu2025mg,yang2024f,chen2025language}.
SOLAMI~\cite{jiang2025solami} advanced this paradigm by enabling interactive dialogue with joint text–motion generation. However, its text-centric alignment strategy neglects the preservation of reasoning capabilities and remains limited in modeling fine-grained synchrony between speech and motion.
In summary, no existing system unifies high-level reasoning, instruction-following, synchronized multimodal generation, and video synthesis within a real-time framework.


To bridge the above limitations, we introduce \textbf{U-Mind}, a unified framework for real-time, high-intelligence multimodal interaction. At its core, U-Mind is built upon a \textit{Unified Alignment and Reasoning Framework} that addresses two challenges: (1) the absence of unified, token-level alignment across modalities, which impairs the temporal and semantic coordination of speech, motion, and language; and (2) the degradation of reasoning capabilities during joint training, which weakens planning and interactive dialogue performance.

To address cross-modal alignment challenges, U-Mind encodes text, audio, and motion into a shared discrete representation space, thus enabling unified multimodal generation in a next-token prediction manner.
A \textit{segment-wise alignment strategy} segments inputs by prosodic boundaries and trains on randomized compositions to further enhance cross-modal temporal synchrony.
To preserve reasoning while enabling grounded multimodal generation, U-Mind adopts a \textit{rehearsal-driven learning strategy}. In the pre-training stage, the model is trained on a carefully balanced mix of modality-aligned tasks, including text-to-motion (T2M), speech-to-motion (S2M), and text-to-speech (T2S), together with large-scale pure-text reasoning data. This joint training simultaneously achieves cross-modal alignment without compromising high-level reasoning ablities.
In the instruction tuning stage, interactive prompts guide the model to generate CoT plans, followed by modality-specific outputs in text, audio, and motion. This text-first decoding strategy ensures that symbolic reasoning and linguistic planning take precedence before generating continuous modalities, thereby better preserving the model’s reasoning capacity throughout the interaction.
To support full-stack multimodal output, we employ a pose-controllable video renderer that synthesizes realistic, temporally aligned videos from generated motion and speech.
Through this design, U-Mind unifies high-level reasoning with continuous multimodal generation within a real-time interaction loop, advancing the development of embodied, instruction-following conversational agents.


In summary, our main contributions are as follows:

\begin{itemize}
   \item We introduce U-Mind, the first unified full-stack multimodal interaction system that supports real-time high-level reasoning dialogue, instruction-following, and generates perceptually complete video responses.

   \item We propose a Unified Alignment and Reasoning Framework that addresses cross-modal synchronization and reasoning degradation through a segment-wise alignment strategy, rehearsal-driven learning, and a text-first decoding strategy.

   \item We demonstrate that U-Mind achieves state-of-the-art performance across a range of multimodal interaction tasks, including high-level question answering, complex instruction execution, and foundational text-to-motion (T2M) and speech-to-motion (S2M) generation.
\end{itemize}

\section{Related Work}
\label{sec:related}
\subsection{Controllable Motion Synthesis}

Early co-speech gesture generation methods relied on rule-based systems with handcrafted templates for speech–motion alignment~\cite{kipp2005gesture, kopp2006towards, cassell1994animated}. With the advent of deep learning, data-driven models~\cite{kucherenko2020gesticulator, yoon2019robots, ahuja2019language2pose} learned direct mappings from speech to motion, significantly improving realism. To better capture the many-to-many nature of gesture semantics, generative approaches~\cite{ginosar2019learning, yoon2020speech, li2021audio2gestures, liu2025tango} introduced motion diversity and naturalness. More recently, diffusion-based models~\cite{zhang2024semantic, zhu2023taming, ao2023gesturediffuclip, deng2025stereo} have shown strong performance in modeling fine-grained dynamics and variability.
The text-to-motion (T2M) task has become a prominent challenge, aiming to generate realistic and contextually appropriate motion from text inputs \cite{ahn2018text2action,barquero2024seamless,chen2023executing,dabral2023mofusion,lu2025scamo,li2025lamp,shen2025bab,newell2025comotion,kim2025controllable,sun2025lal,zhou2025modeseq,wu2025moner,zhang2025monst3r,huang2025move,curreli2025nonisotropic,kim2025personabooth,meng2025rethinking}. Recent advances in generative models, particularly those utilizing diffusion techniques \cite{hong2025salad,hua2025deterministic,cong2025semgeomo,li2025simmotionedit,ji2025towards}, have led to more robust T2M pipelines capable of handling complex motion dynamics \cite{tian2025direct,matada2025generalizable,zhao2025dartcontrol,li2025latenthoi,ait2025learning,ruiz2025mixermdm,ji2025pomp,jiang2024scaling,liao2025shape,wang2025stickmotion,shin2024wham}.
More recently, large language models (LLMs) have been integrated into motion synthesis pipelines to enhance contextual understanding and enable instruction-aware generation \cite{jiang2023motiongpt,wu2025mg,yang2024f,wu2025motion,han2025atom,cong2025semgeomo,tan2025think,zhao2025efficient,pan2025chain,li2025lamp,wang2024move,lu2025scamo,zhang2025social,li2024cposer}, particularly in multimodal interactive systems.
Despite these advances, existing approaches typically focus on isolated generation tasks and lack unified, reasoning-aware integration across modalities. Achieving coherent, real-time generation of speech, text, and motion with high-level understanding remains an open challenge.


\begin{figure*}
    \centering
    \includegraphics[width = \textwidth]{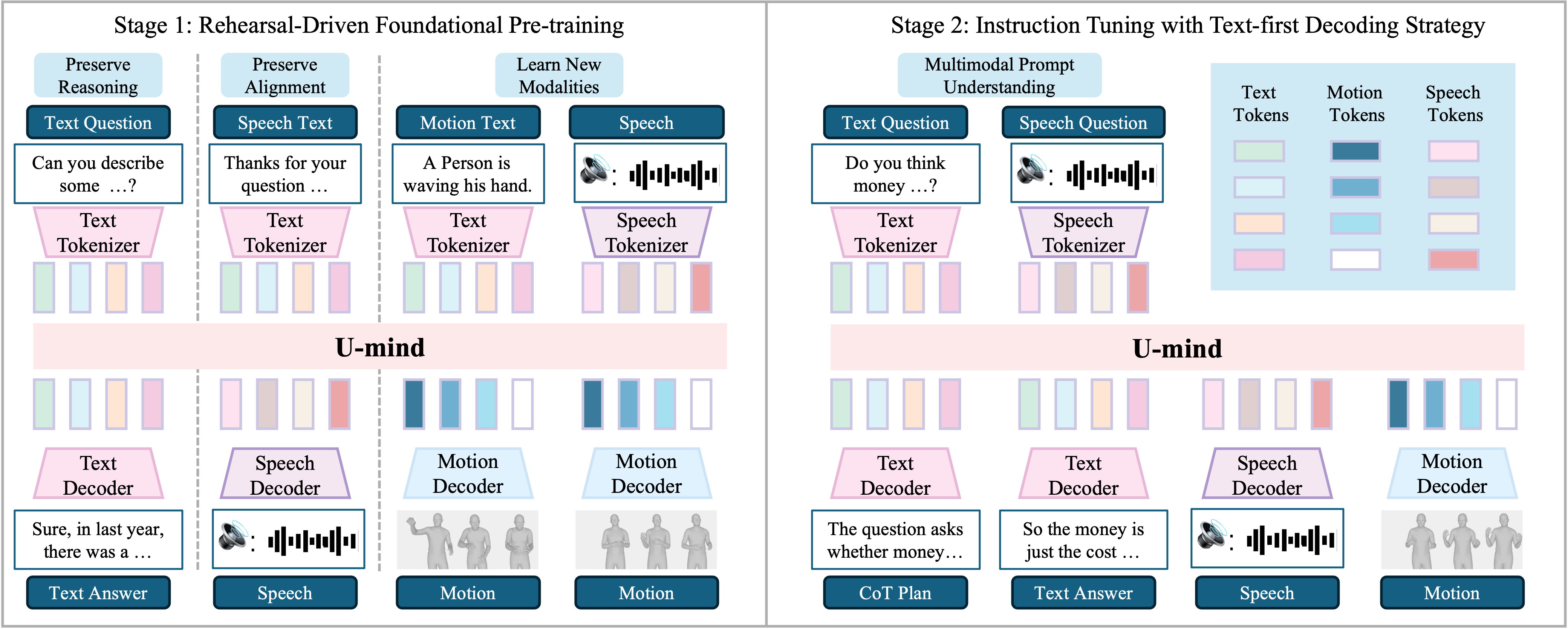}
    \caption{Method Overview.
Our framework adopts a two-stage training paradigm. In Stage 1, we conduct rehearsal-driven pretraining to preserve symbolic reasoning (Textual QA), maintain speech alignment (Text2Speech), and learn new modalities (Text2Motion, Speech2Motion). All tasks are unified via discrete tokens processed by a shared U-mind backbone. In Stage 2, we instruction-tune the model with multimodal prompts (text or audio), generating CoT plans followed by coherent outputs across modalities.
}
    \label{fig:method}
\end{figure*}

\subsection{Embodied Intelligence}

Embodied intelligence has evolved from simple avatars with scripted responses and rule-based non-verbal cues to advanced systems that integrate multimodal perception and LLMs for reasoning~\cite{liu2025aligning, duan2022survey, fu2025ins, javed2025intermask, fang2025llama, wu2024next}. Early work focused on template-driven dialogues and handcrafted gestures~\cite{bickmore2005social, cassell2001embodied, cassell2001non}. Recent efforts explore neural architectures that jointly process speech, vision, and text~\cite{deichler2023diffusion, jang2024faces, fan2024freemotion, yao2024moconvq, wei2025motion, hong2025motionbench, shao2025ulmulti, xia2025phoenix, zhen2025teller, defossez2024moshi}. More recently, LLM-based systems unify open-domain dialogue and motion generation within a single framework~\cite{chen2025language, jiang2023motiongpt, jiang2025solami, wang2025aligning, wang2025timotion, xie2025x, grauman2024ego, huang2024embodied}. However, these approaches typically decouple reasoning, dialogue, and realistic response, lacking real-time conversational interaction with tightly synchronized multimodal output. 


\section{Method}
\label{sec:method}
In this section, we present \textbf{U-Mind}, a unified framework for real-time, high-intelligence multimodal interaction. 
We begin by outlining the overall system architecture (Sec.~\ref{sec:method:overview}) and the unified representation space along with core network design (Sec.~\ref{sec:method:arch}). We then detail the two-stage training process: Rehearsal-Driven Foundational Pre-training (Sec.~\ref{sec:method:stage1}), which addresses the challenge of reasoning degradation during multimodal alignment, and Instruction Tuning with Text-First Decoding Strategy (Sec.~\ref{sec:method:stage2}), which prepares the model for text-first decoding pipeline and instruction-following. Finally, we describe the real-time interaction pipeline (Sec.~\ref{sec:method:inference}) that supports closed-loop, multi-turn dialogue with synchronized outputs across text, audio, motion, and video.


\subsection{System Overview}
\label{sec:method:overview}

A straightforward strategy for building a real-time multimodal agent might involve directly fine-tuning a pre-trained large language models (LLMs) on a mixture of multimodal instruction data. However, such one-stage supervised fine-tuning (SFT) often results in obviously modality misalignment~\cite{jiang2025solami,yang2024f}. Introducing perceptual modalities such as motion and speech creates competing objectives: low-level adaptation versus high-level reasoning. These conflicts often lead to catastrophic forgetting of the LLM’s original planning and dialogue capabilities.

To address this, we design a \textit{Unified Alignment and Reasoning Framework} that explicitly addresses the challenges of multimodal alignment and high-level reasoning degredation. In pre-training stage, the model is exposed to a carefully balanced mixture of modality-aligned supervision: text-to-speech (T2S), text-to-motion (T2M), and speech-to-motion (S2M), alongside pure-text “rehearsal” data. This stage enables the model to acquire new modality fluency while retaining its core reasoning capability. In instruction tuning stage, the model is aligned to human intent through a diverse corpus of CoT-driven prompts covering intelligent dialogue and instruction-following.
This two-stage process equips U-Mind with both unified cross-modal alignment and preserved high-level reasoning, enabling real-time multimodal dialogue and embodied interaction.
The overview of our method is shown in Fig \ref{fig:method}.

\subsection{Architecture and Modality Representation}
\label{sec:method:arch}

Our system is built upon the LLaMA2-7B \cite{touvron2023llama} large language model backbone. To extend this text-only foundation into a unified multimodal framework, we following \cite{zhan2024anygpt, jiang2025solami} to introduce discrete token representations for speech and body motion, mapping all modalities into a shared embedding space.

\noindent\textbf{Motion Representation.}  
We represent human motion using the SMPL-X body model \cite{SMPL-X:2019}. Pose parameters are first converted into continuous 6D joint rotations, which are more stable and expressive than Euler angles or quaternions. To bridge this continuous representation with the LLM’s discrete token space, we train a Residual Vector Quantized VAE (RVQ-VAE) \cite{siyaoduolando} on the 6D pose sequences. This module discretizes motion into sequences of latent motion tokens, enabling autoregressive modeling over temporally structured body dynamics.


\noindent\textbf{Speech Representation.}  
For the speech modality, we adopt an analogous discretization strategy using the RVQ-VAE architecture from SpeechTokenizer \cite{zhang2023speechtokenizer}, a state-of-the-art neural audio encoder. This model compresses raw waveforms into discrete acoustic tokens that capture both semantic content and paralinguistic cues (e.g., prosody, emotion), facilitating symbolic-level processing of audio streams.

\noindent\textbf{Special Reasoning Tokens.}  
To explicitly support high-level reasoning, we introduce two special tokens: $\langle\text{think}\rangle$ and $\langle/\text{think}\rangle$. These delimit internal \textit{Chain-of-Thought} (CoT) segments within each response, allowing the model to articulate intermediate reasoning steps before generating observable outputs. Content within these tokens is text-only and used solely for internal planning.
For modality structuring, speech, motion, and reasoning segments are wrapped with start and end tokens, while plain text is generated without explicit delimiters. A global start and end token further encapsulate the full response sequence.

\noindent\textbf{Unified Embedding Space.}  
To support multimodal token modeling, we resize the LLM’s vocabulary and embedding matrix to incorporate the motion and speech token sets, along with the CoT reasoning tokens. This yields a unified token space where all modalities (text, speech, motion, and reasoning) are jointly represented. 
Consequently, the model operates autoregressively, generating interleaved multimodal sequences via next-token prediction.

\subsection{Rehearsal-Driven Foundational Pre-training}
\label{sec:method:stage1}

The goal of this foundational stage is to enable the model to generate and understand new modalities without compromising its pre-trained reasoning capabilities. To achieve this, we adopt a \textit{Rehearsal-Driven Learning} strategy that combines perceptual grounding with symbolic rehearsal.
Specifically, we construct a carefully curated training mixture consisting of two task types:
(1) \textit{Modality grounding tasks}, including text-to-motion (T2M), speech-to-motion (S2M), and text-to-speech (T2S), which teach the model to produce temporally coherent and contextually appropriate sequences from linguistic or acoustic prompts.
To further enhance the alignment between modalities, we employ a \textit{segment-wise alignment strategy}, which segments inputs based on rhythm and pauses, and trains the model on randomized combinations of these segments. This strategy facilitates fine-grained temporal alignment and cross-modal correspondence learning.
(2) \textit{Rehearsal tasks}, which include high-quality pure-text reasoning data to preserve the LLM’s planning and dialogue abilities. This mixture ensures the model to acquire grounded multimodal generation skills while continuously reinforcing its symbolic reasoning core. By jointly optimizing these competing objectives in a balanced manner, we effectively mitigate reasoning degradation while achieving robust cross-modal alignment. The result is a stable, high-capacity backbone that serves as the foundation for downstream instruction tuning.

\subsection{Instruction Tuning with Text-first Decoding Strategy}
\label{sec:method:stage2}

Building on the high-intelligence foundation established in pre-training stage, the second stage aims to align the model with complex user instructions and multimodal interaction behaviors. To this end, we perform supervised fine-tuning on a instruction-following corpus comprising diverse dialogue and task prompts. 
A central component of this stage is our \textit{text-first decoding strategy}. Each model response begins with an internal reasoning block enclosed by $\langle\text{think}\rangle$ and $\langle/\text{think}\rangle$, which serves as a latent CoT-style plan. This plan guides the subsequent generation of coherent text, speech, and motion outputs.
This formulation unifies complex instruction-following and conversational question answering into a single prompt-response interface. The model learns to interpret intent, perform intermediate reasoning, and produce coherent multimodal behavior, all within an autoregressive decoding pipeline.

\subsection{Inference Pipeline for Real-time Interaction}
\label{sec:method:inference}

Given a user prompt, U-Mind performs autoregressive decoding to produce a structured output sequence: (1) an internal CoT plan enclosed by $\langle\text{think}\rangle$ tags, (2) a textual response, (3) acoustic tokens, and (4) motion tokens. These components are temporally aligned to form a coherent multimodal response.
For video synthesis, we support two rendering backends:
(1) a diffusion-based renderer that synthesizes photorealistic 2D videos, conditioned on 2D keypoints projected from SMPL-X poses using DWPose~\cite{yang2023effective};
(2) a Gaussian Splatting renderer that directly renders 3D human videos from SMPL-X pose sequences.
This dual-path rendering design enables full-stack generation from user inputs to synchronized text, audio, motion, and video.

\section{Experiments}
\label{sec:experiments}

\subsection{Experimental Settings}
\noindent\textbf{Datasets.}
We conduct training and evaluation on two benchmark datasets: BEAT v2~\cite{liu2024emage} for speech-to-motion (S2M) and HumanML3D~\cite{Guo_2022_CVPR} for text-to-motion (T2M). To inject explicit reasoning signals, we augment each sample with three QA-style triplets generated via Qwen3~\cite{qwen3}, resulting in 10,000 S2M and 16,000 T2M sentence-level samples. Each dataset is split into training and testing sets with a 7:1 ratio.
To preserve language reasoning ability, we additionally include OpenOrca~\cite{mukherjee2023orca}, a high-quality corpus of conversational CoT data. The TTS module is trained on Common Voice~\cite{ardila2020common}, a large-scale multilingual speech dataset that enhances prosody and phonetic coverage.
For video rendering, we collect 400 hours of proprietary human recordings, annotated with 2D keypoints using DWPose~\cite{yang2023effective}, enabling high-fidelity motion synthesis synchronized with generated dialogue and instructions.


\noindent\textbf{Baselines.}  
We evaluate our model against several representative baselines, including both interactive multimodal agents and unimodal synthesis models:

\begin{itemize}
    \item \textbf{SOLAMI~\cite{jiang2025solami}.} An end-to-end system for multimodal dialogue with speech and motion generation. It serves as the primary baseline for evaluating real-time interactive performance.

    \item \textbf{LOM~\cite{chen2025language}.} A state-of-the-art model for text-to-motion (T2M) and speech-to-motion (S2M) tasks, lacking reasoning or dialogue capabilities. To further evaluate dialogue-level performance, we additionally construct a pipeline baseline combining LLaMA2-7B-chat \cite{touvron2023llama}, Orpheus-TTS, and LOM for speech and motion synthesis conditioned on generated responses.

    
    \item \textbf{EMAGE~\cite{liu2024emage}.} A speech-to-motion model that leverages masked body gesture priors during training to enhance inference. It integrates a Masked Audio Gesture Transformer for joint audio-to-gesture generation and masked gesture reconstruction.
    
    \item \textbf{CaMN~\cite{liu2022beat}.} A speech-to-motion model that employs a cascaded architecture for gesture synthesis, enabling the generation of complex body motions from speech.

    \item \textbf{DisCo~\cite{liu2022disco}.} A speech-to-motion model that disentangles content and rhythm features via contrastive learning, and employs a Diversity-and-Inclusion Network to improve motion diversity and mitigate mode collapse.


\end{itemize}


\begin{figure*}
    \centering
    \includegraphics[width = 0.8\textwidth]{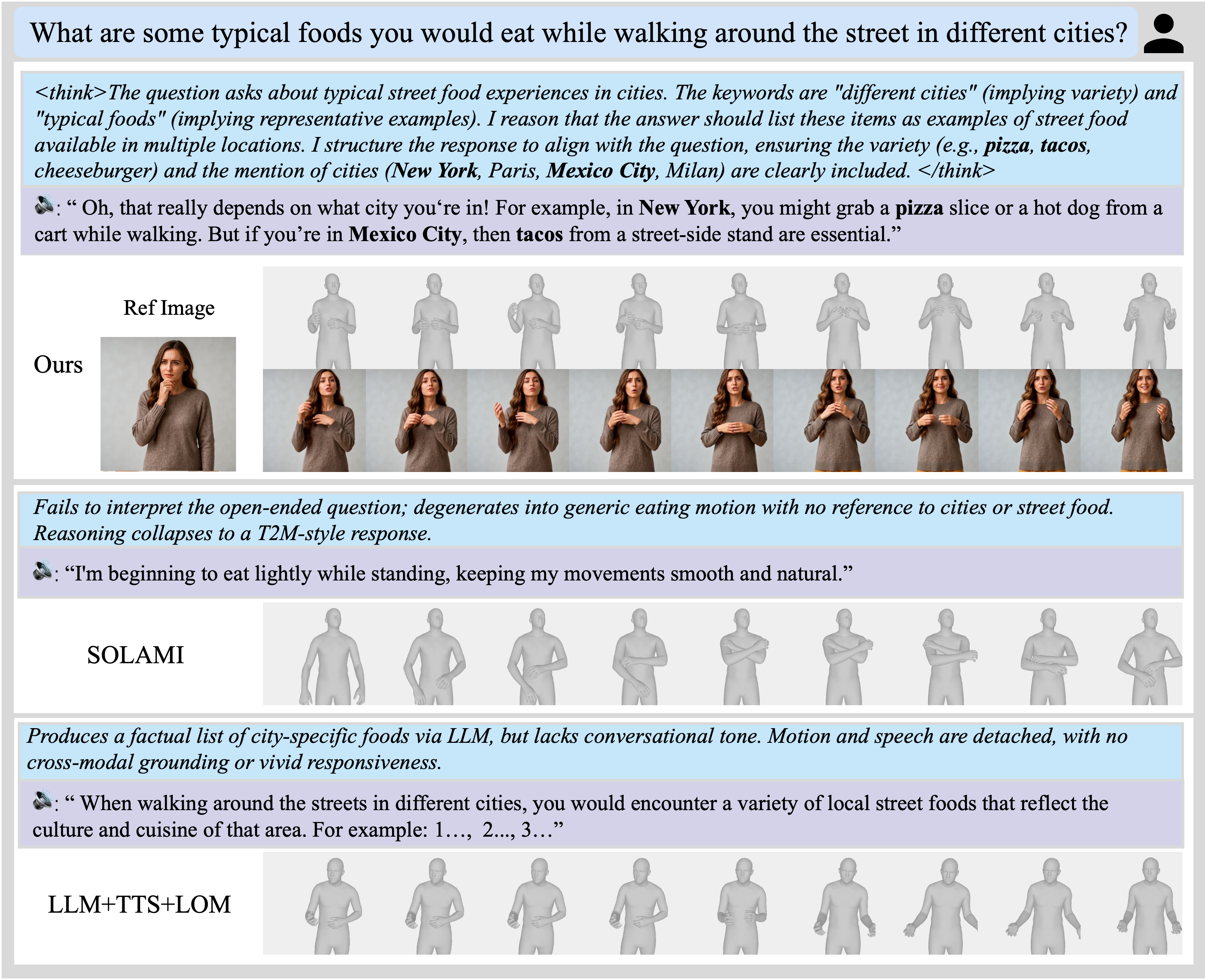}
    \caption{Multimodal Dialogue Results. U-Mind performs CoT-based reasoning and generates synchronized speech and motion, producing photorealistic, context-aware responses. In contrast, SOLAMI degenerates into generic gestures without understanding the prompt, while LLM+TTS+LOM lacks coherence and cross-modal grounding.}
    \label{fig:dialogue_results}
\end{figure*}

\noindent\textbf{Implementation Details.}
Our core model is built upon the AnyGPT (LLaMA2-7B) architecture~\cite{zhan2024anygpt}. Pre-training stage was performed on 8 H100 GPUs using the AdamW optimizer~\cite{loshchilovdecoupled}, with a peak learning rate of $1 \times 10^{-4}$ and cosine decay. For instruction tuning stage, we employed the same setup but reduced the learning rate to $2 \times 10^{-5}$ to stabilize alignment. The WAN-based video generation module~\cite{wan2025wan} was fine-tuned on 16 H100 GPUs using AdamW with a learning rate of $1 \times 10^{-5}$.
Speech audio was transcribed using Whisper \cite{whisper2022}
, with segmentation at punctuation and pause boundaries. Synthetic speech for the speech-to-motion (S2M) task was generated using Orpheus-TTS \cite{orpheus2025}
. Motion data was tokenized using our RVQ-VAE~\cite{siyaoduolando} and decoded into 6D poses. 

\noindent\textbf{Evaluation Metrics.} 
We evaluate our multimodal interaction system along two key dimensions: motion quality and reasoning ability.
We adopt both distributional and pose-level metrics. \textbf{Fréchet Gesture Distance (FGD)}~\cite{yoon2020speech} measures the realism and distributional alignment of generated gestures with ground-truth motion. \textbf{Diversity}~\cite{li2021audio2gestures} quantifies the expressiveness and variability of the generated motion sequences. Additionally, we compute the \textbf{Angle Error} \cite{hong2025egolm}, defined as the mean angular deviation between predicted and reference body joint rotations, to assess low-level pose accuracy.
To evaluate the quality of generated responses, we employ a large language model (Qwen) as an automatic judge. It assigns 10-point scores along two axes: \textbf{Relevance}, which measures how well the response addresses the user's intent and topical focus, and \textbf{Naturalness}, which reflects fluency and human-likeness in language generation.
Together, these metrics provide a rigorous and multidimensional assessment of both the perceptual and cognitive capabilities of the system.

\begin{figure*}
    \centering
    \includegraphics[width = 0.8\textwidth]{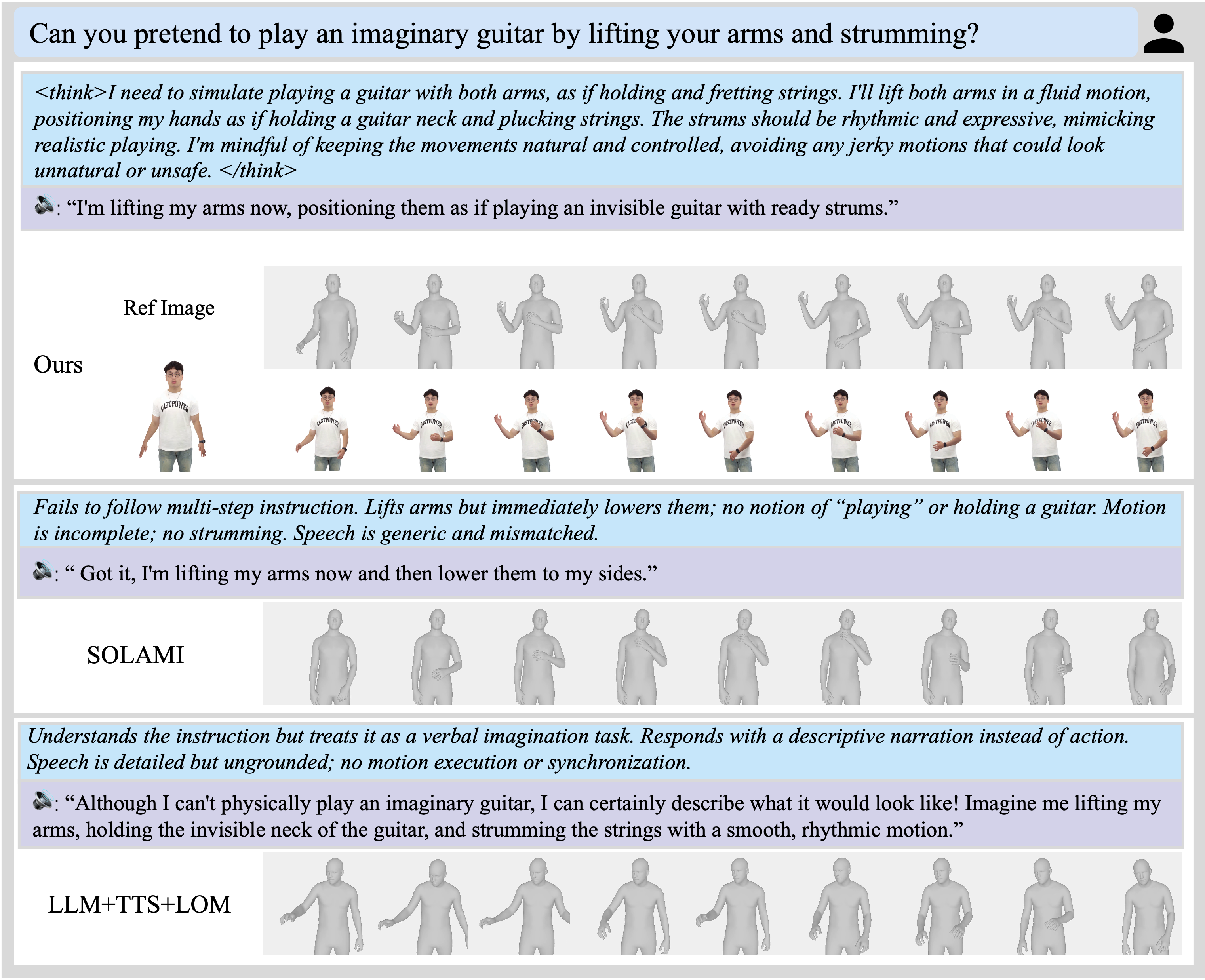}
    \caption{Intruction-following Results. U-Mind interprets the user's intent through CoT planning and generates expressive, context-aware motions with realistic video output.
In contrast, SOLAMI produces a shallow, literal response without understanding or simulating the intended imaginary action, while LLM+TTS+LOM lacks embodiment and cross-modal synchronization.}
    \label{fig:instruct_results}
\end{figure*}

\begin{table}
\centering 
\caption{Multimodal Dialogue Results. The \textbf{bolded} numbers represent the best results, while the \underline{underlined} numbers indicate the second-best results.
}
\label{tab:dialogue_results} 
\begin{adjustbox}{width=0.47\textwidth}
\setlength{\tabcolsep}{5pt}
\begin{tabular}{ccccc}
\toprule 
Methods & \textbf{FGD} $\downarrow$ & \textbf{Diversity} $\uparrow$ & \textbf{Relevance} $\uparrow$ & \textbf{Naturalness} $\uparrow$ \\
\midrule
Dataset & 0 & 11.37 & 8.32 & 8.57\\
\midrule
LLM+TTS+LOM \cite{chen2025language} & \underline{17.87} & \underline{11.02} & \textbf{8.72} & 3.95\\
SOLAMI \cite{jiang2025solami} & 18.43 & 9.29 & 1.23 & \underline{5.62}\\
Ours & \textbf{7.67} & \textbf{11.18} & \underline{8.23} & \textbf{8.11}\\
\bottomrule 
\end{tabular}
\end{adjustbox}
\end{table}

\subsection{Results and Comparisons}

We conduct comprehensive experiments to evaluate the performance of our model across two primary dimensions. First, we assess its effectiveness in high-level interaction tasks such as multimodal dialogue and instruction following, which constitute the central focus of this work. Second, we examine its ability to preserve generation quality in foundational tasks like speech-to-motion (S2M) and text-to-motion (T2M), thereby validating the robustness of our training framework.

\begin{table}
\centering 
\caption{Instruct Following Results. The \textbf{bolded} numbers represent the best results, while the \underline{underlined} numbers indicate the second-best results.
}
\label{tab:instruct_results} 
\begin{adjustbox}{width=0.47\textwidth}
\setlength{\tabcolsep}{5pt}
\begin{tabular}{ccccc}
\toprule 
Methods & \textbf{FGD} $\downarrow$ & \textbf{Diversity} $\uparrow$ & \textbf{Relevance} $\uparrow$ & \textbf{Naturalness} $\uparrow$ \\
\midrule
Dataset & 0 & 10.56 & 9.17 & 8.68\\
\midrule
LLM+TTS+LOM \cite{chen2025language} & \underline{10.73} & 7.96 & \textbf{9.00} & 6.26\\
SOLAMI \cite{jiang2025solami} & 18.51 & \underline{10.01} & 7.56 & \underline{7.92}\\
Ours & \textbf{5.12} & \textbf{10.19} & \underline{8.50} & \textbf{8.26}\\
\bottomrule 
\end{tabular}
\end{adjustbox}
\end{table}

\begin{table}[t] 
\centering 
\caption{
    Quantitative comparisons for Speech-to-Motion (S2M) synthesis.
    The \textbf{bolded} numbers represent the best results, while the \underline{underlined} numbers indicate the second-best results.
}
\label{tab:s2m_results} 
\begin{adjustbox}{width=0.4\textwidth}
\begin{tabular}{lccc} 
\toprule
Methods & \textbf{FGD} $\downarrow$ & \textbf{Angle Error} $\downarrow$ & \textbf{Diversity} $\uparrow$ \\
\midrule
Dataset & 0 & 0 & 10.88 \\
\midrule
Camn \cite{liu2022beat} & 20.37 & 0.264 & 8.85\\
Disco \cite{liu2022disco} & 20.32 & 0.260 & 10.80 \\
EMAGE  \cite{liu2024emage}  & 17.85 & \underline{0.248} & 11.20 \\
LOM \cite{chen2025language}    & \underline{16.47} & 0.251 & \textbf{11.96} \\
Ours        & \textbf{11.12} & \textbf{0.188} & \underline{11.48} \\
\bottomrule
\end{tabular}
\end{adjustbox}
\end{table}
\begin{table}[t] 
\centering 
\caption{
    Quantitative comparisons for Text-to-Motion (T2M) synthesis.
    The \textbf{bolded} numbers represent the best results, while the \underline{underlined} numbers indicate the second-best results.
}
\label{tab:t2m_results} 
\begin{adjustbox}{width=0.4\textwidth}
\begin{tabular}{lccc} 
\toprule
Methods & \textbf{FGD} $\downarrow$ & \textbf{Angle Error} $\downarrow$ & \textbf{Diversity} $\uparrow$ \\
\midrule
Dataset & 0 & 0 & 11.48 \\
\midrule
LOM \cite{chen2025language}    & 14.22 & \underline{0.331} & \underline{10.42} \\
SOLAMI \cite{jiang2025solami}    & \textbf{8.64} & 0.336 & 7.36 \\
Ours         & \underline{12.69} & \textbf{0.109} & \textbf{10.71}\\
\bottomrule
\end{tabular}
\end{adjustbox}
\vspace{-3mm}
\end{table}

\noindent\textbf{Performance in High-Level Interaction Tasks.} 
We evaluate high-level interaction capabilities through two representative tasks: \textit{multimodal dialogue} and \textit{instruction following}, both of which demand coherent response generation, context-aware reasoning, and synchronized multimodal outputs.
As shown in Table~\ref{tab:dialogue_results}, U-Mind achieves the best overall performance in the dialogue setting, outperforming all baselines in motion quality, generation diversity, and naturalness. While the LLM+TTS+LOM pipeline achieves slightly higher relevance scores due to its reliance on a strong language model, it suffers from low naturalness and weak cross-modal coordination, resulting in flat and less vivid interactions. In contrast, U-Mind produces more fluid and perceptually grounded interactions through unified multimodal generation.
In the instruction-following task (Table~\ref{tab:instruct_results}), U-Mind again leads in most metrics, demonstrating its ability to handle complex, goal-directed interaction. Compared to prior systems, it offers more expressive and temporally aligned responses, benefiting from our text-first decoding and alignment strategy. 
Visualizations in Figures~\ref{fig:dialogue_results} and \ref{fig:instruct_results} further highlight U-Mind’s advantages in producing synchronized and semantically coherent outputs, validating the effectiveness of our \textit{Unified Alignment and Reasoning Framework} in enabling high-intelligence multimodal interaction.
\noindent\textbf{Performance in Foundational Synthesis Tasks.}
We further evaluate the model's ability to generate high-quality motion in low-level synthesis settings, namely text-to-motion (T2M) and speech-to-motion (S2M). These tasks validate whether our unified training pipeline maintains motion fidelity and diversity across modalities.
As shown in Table~\ref{tab:t2m_results}, U-Mind achieves the best diversity and lowest angle error in the T2M task, indicating high expressiveness and precise motion dynamics. Although SOLAMI reports a lower FGD, this result is less indicative of general performance, as SOLAMI is specified to T2M task and thus benefits from task-specific optimization without the challenges of cross-modal integration. In contrast, U-Mind maintains competitive FGD while supporting joint T2M and S2M generation, demonstrating the effectiveness of our unified training strategy in preserving motion quality across modalities.
Table~\ref{tab:s2m_results} presents the results for S2M generation. U-Mind outperforms all baselines in both FGD and angle error, demonstrating superior temporal fidelity and motion smoothness. While LOM slightly leads in diversity, our model achieves a comparable score while significantly improving realism, as evidenced by a notable drop in FGD and angle error. These results confirm that our model maintains strong low-level synthesis capabilities, even when jointly trained with high-level reasoning and dialogue objectives.

\subsection{Ablation Studies}

\noindent\textbf{Impact on Reasoning and Generation Quality.} 
We perform ablation studies to quantify the contribution of each reasoning-related component in our framework. As shown in Table~\ref{tab:abalation}, removing data rehearsal during pre-training leads to noticeable drops in both relevance and naturalness, confirming the importance of symbolic pretraining in preserving high-level reasoning during multimodal generation.
Removing text-first decoding, where the model directly generates response speech and motion without intermediate text, causes a sharp decline in relevance. This suggests that text-first decoding serves as a critical planning scaffold, aligning content generation across modalities and maintaining semantic integrity.
Disabling chain-of-thought (CoT) reasoning during instruction tuning also results in reduced relevance, demonstrating that explicit reasoning traces help structure response generation and improve task understanding. These results validate the effectiveness of our method in preserving reasoning ability and enhancing the coherence of multimodal interaction.

\noindent\textbf{Impact of Segment-Wise Alignment.} 
We investigate the effect of our segment-wise alignment strategy by comparing it with a variant trained on full utterances without segmentation. As shown in Table~\ref{tab:abalation_s2m}, removing segmentation results in degraded motion quality across all metrics. These results highlight the benefit of segment-based training in capturing fine-grained rhythm and improving temporal synchronization between speech and motion. By leveraging rhythmic boundaries and random segment combinations, our method encourages better cross-modal alignment and generates more expressive, natural gestures.






\begin{table}[t] 
\centering 
\caption{
    Quantitative comparisons for Ablation studies.
}
\label{tab:abalation} 
\begin{adjustbox}{width=0.4\textwidth}
\begin{tabular}{lcc} 
\toprule
Methods & \textbf{Relevance} $\uparrow$& \textbf{Naturalness}  $\uparrow$\\
\midrule
wo-data rehearsal    & 6.13 & 7.18  \\
wo-text-first    & 1.24 & 5.18  \\
wo-cot    & 5.54 & 7.23  \\
\midrule
Ours        & 8.23 & 8.11 \\
\bottomrule
\end{tabular}
\end{adjustbox}
\end{table}

\begin{table}[t] 
\centering 
\caption{
    Quantitative comparisons for Ablation studies.
}
\label{tab:abalation_s2m} 
\begin{adjustbox}{width=0.4\textwidth}
\begin{tabular}{lccc} 
\toprule
Methods & \textbf{FGD} $\downarrow$ & \textbf{Angle Error} $\downarrow$ & \textbf{Diversity} $\uparrow$  \\
\midrule
wo-seg    & 16.89 & 0.219 & 10.46  \\
\midrule
Ours        & 11.12 & 0.188 & 11.48 \\
\bottomrule
\end{tabular}
\end{adjustbox}
\end{table}

\section{Discussion and Conclusion} 
\label{sec:conclusion}

\noindent\textbf{Limitations.}  
While our work establishes a robust paradigm for real-time multimodal interaction, several limitations remain.  
First, the expressiveness of generated motion is constrained by the discrete vocabulary of the motion quantizer, limiting the fidelity of fine-grained gestures such as facial expressions and subtle hand movements.  
Second, the data composition used in pretraining was determined empirically. A more principled framework is needed to balance symbolic rehearsal and new modality acquisition, potentially improving the generalization and stability of multitask learning.

\noindent\textbf{Social Impact.}  
Interactive digital humans present a dual-use challenge: they enable valuable applications in accessibility, education, and entertainment, but also risk misuse, such as deepfakes for misinformation or fraud. We advocate for parallel efforts in synthetic media detection and data de-biasing to mitigate these risks.

\noindent\textbf{Conclusion.}  
We present U-Mind, the first unified, real-time, and full-stack multimodal interaction system that supports high-level dialogue, instruction following, and perceptually grounded video generation. At its core, U-Mind implements a \textit{Unified Alignment and Reasoning Framework}, which integrates segment-wise alignment strategy, rehearsal-driven learning, and text-first decoding strategy to jointly preserve reasoning ability and enhance cross-modal synchronization.
Comprehensive experiments demonstrate that U-Mind achieves state-of-the-art performance across both high-level interaction tasks and low-level generation tasks.
By unifying high-level reasoning with real-time multimodal generation within a single interaction loop, U-Mind marks a significant step toward building highly intelligent and immersive interactive agents.


{
    \small
    \bibliographystyle{ieeenat_fullname}
    \bibliography{main}
}


\end{document}